\documentclass[conference]{IEEEtran}
\usepackage{times}

\usepackage[numbers]{natbib}
\usepackage{multicol}
\usepackage[bookmarks=true,backref=page]{hyperref}
\usepackage{graphicx}
\usepackage{booktabs}
\usepackage{subcaption}
\usepackage{multirow}
\usepackage{algorithm}
\usepackage{algorithmic}
\usepackage{amsmath}
\usepackage{amsfonts}
\usepackage[dvipsnames]{xcolor}
\usepackage{flushend}
\usepackage{makecell}

\definecolor{talents}{RGB}{3,169,244}

\begin{document}

\title{Modeling Latent Partner Strategies\\for Adaptive Zero-Shot Human-Agent Collaboration}

\author{\authorblockN{Benjamin Li\authorrefmark{1},
Shuyang Shi\authorrefmark{1},
Lucia Romero\authorrefmark{2}, 
Huao Li\authorrefmark{2},
Yaqi Xie\authorrefmark{1},
Woojun Kim\authorrefmark{1}, \\
Stefanos Nikolaidis\authorrefmark{3},
Michael Lewis\authorrefmark{2},
Katia Sycara\authorrefmark{1}, and
Simon Stepputtis\authorrefmark{1}\authorrefmark{4}
}
\authorblockA{\authorrefmark{1}
Carnegie Mellon University,
\authorrefmark{2}
University of Pittsburgh,
\authorrefmark{3}
University of Southern California
\authorrefmark{4}
Virginia Tech
}
}

\maketitle

\begin{abstract}
In collaborative tasks, being able to adapt to your teammates is a necessary requirement for success. 
When teammates are heterogeneous, such as in human-agent teams, agents need to be able to observe, recognize, and adapt to their human partners in real time.
This becomes particularly challenging in tasks with time pressure and complex strategic spaces where the dynamics can change rapidly.
In this work, we introduce TALENTS, a strategy-conditioned cooperator framework that learns to represent, categorize, and adapt to a range of partner strategies, enabling ad-hoc teamwork.
Our approach utilizes a variational autoencoder to learn a latent strategy space from trajectory data. This latent space represents the underlying strategies that agents employ. Subsequently, the system identifies different types of strategy by clustering the data. Finally, a cooperator agent is trained to generate partners for each type of strategy, conditioned on these clusters. 
In order to adapt to previously unseen partners, we leverage a fixed-share regret minimization algorithm that infers and adjusts the estimated partner strategy dynamically.
We assess our approach in a customized version of the Overcooked environment, posing a challenging cooperative cooking task that demands strong coordination across a wide range of possible strategies.
Using an online user study, we show that our agent outperforms current baselines when working with unfamiliar human partners.
\end{abstract}

\IEEEpeerreviewmaketitle

\section{Introduction}

As AI agents and robots become increasingly integrated into daily life, the development of methods for effective human-agent collaboration is more critical than ever. 
In ad hoc teamwork settings, where agents must cooperate without prior knowledge of the partner, success depends on accurately predicting partner behavior and selecting actions that maximize joint performance~\cite{stone2010ad}. 
To be effective, collaborative agents must interpret the behaviors of their partners in real time and respond quickly and effectively to account for the diverse, non-stationary, and often suboptimal strategies employed by human partners.

\begin{figure}
    \centering
    \includegraphics[width=0.8\linewidth]{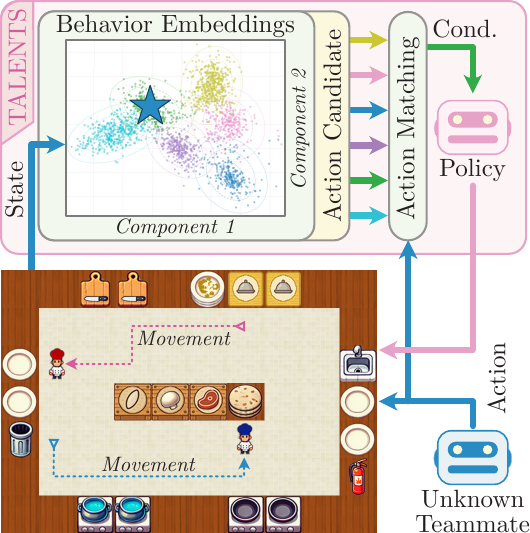}
    \caption{The TALENTS framework: Given an unknown teammate’s observation, strategy clusters are used to generate action predictions. These predictions are then used to infer teammate type and corresponding best responses using a tracking-regret minimization algorithm.}
    \label{fig:overview}
\end{figure}

To this end, we introduce \textbf{T}eam \textbf{A}daptation via \textbf{L}at\textbf{E}nt \textbf{N}o-regre\textbf{T} \textbf{S}trategies (\textbf{TALENTS}) (see Fig.~\ref{fig:overview}), a novel method for zero-shot coordination that models partner behaviors in a semantic strategy space.
Our agent utilizes a latent behavior representation learned using a variational autoencoder (VAE).
The VAE is trained offline on trajectory data collected from a diverse population of baseline agents, trained with various reward-shaped targets, with the objective of predicting each agent’s next high-level action (e.g. ``pick up a plate'').
Subsequently, the latent space is utilized to capture meaningful behavioral patterns through K-Means clustering, using silhouette analysis~\citep{rousseeuw1987silhouettes} to determine the optimal number and placement of such clusters.
The TALENTS cooperator training process leverages the VAE's generative capability to produce partner agents by sampling from the clusters, while conditioning the cooperator on those clusters to develop strategy-specific responses.
At test time, the agent infers the type of a new partner by comparing a sampled action from each behavior cluster with the partner's actually observed action, utilizing the cluster with the best-matching action as likely partner type.
TALENTS continuously updates its belief and conditions its own actions on the inferred partner type to enable more effective coordination, thereby behaving like an optimal response to the estimated partner.

To test our agent, we utilize a modified Overcooked setting with a set of human subjects in an ad-hoc teamwork task. 
In contrast to prior Overcooked environments, our setting introduces additional temporal pressure through timers on orders, as well as bonus rewards for fast deliveries. Moreover, it requires the utilization of three cooking stations for two distinct recipes, significantly increasing the need for effective teamwork.
In summary, we present a novel approach for strategy-conditioned human-agent collaboration, enabling our agent to infer teammate types and adapt its behavior in real time, achieving state-of-the-art zero-shot performance in a challenging Overcooked evaluation with previously unseen human partners.
Furthermore, we demonstrate that human partners prefer to interact with our agent based on the perceived fluency and trustworthiness of their teamwork, further underscoring the effectiveness of our adaptive agent.

\section{Related Work}
\paragraph{Strategy Inference} 
The ability to infer a teammate’s strategy during collaboration enables agents to coordinate effectively.
Methods for strategy inference include learning latent embeddings from trajectory datasets~\citep{nikolaidis2015efficient,zhao2022coordination,xie2021learning, hong2023learning}, estimating policy updates~\citep{foerster2018learning}, and Bayesian inference~\citep{chalkiadakis2003coordination,foerster2019bayesian, zintgraf2021deep}. Other approaches use triplet losses for distinguishability~\citep{grover2018learning}, action prediction~\citep{papoudakis2020variational, raileanu2018modeling}, or maintaining strategy libraries~\citep{9540646, he2016opponent}. Theory of Mind modeling is also often used as a technique for predicting teammate intentions and beliefs~\citep{9900572,rabinowitz2018machine, oguntolatheory, li2023theory}. TALENTS learns latent representations of teammate strategies using a variational autoencoder, unsupervised clustering to identify similar characteristics, then performs partner inference via tracking-regret minimization at test-time settings to adapt to teammates.

\paragraph{Zero-Shot Coordination} 
The central aim of Zero-Shot Coordination (ZSC) is to develop agents that are able to collaborate with partners that were unseen to the agent during train time~\citep{hu2020other}. Two common methods to achieve this are Self-play (SP)~\citep{5392560,heinrich2015fictitious} and Population-Based Training (PBT)~\citep{jaderberg2019human}. SP methods have shown remarkable performance in zero-sum environments~\citep{silver2017mastering,vinyals2019grandmaster}, but are unable to achieve comparable performance in common-payoff cooperative settings since self-play agents tend to develop rigid behavior patterns~\citep{carroll2019utility}. PBT encourages less rigid convention formation by optimizing over diverse partner sets~\citep{lupu2021trajectory,strouse2021collaborating,charakorn2023generating, zhao2023maximum}, enabling robustness to a greater variety of partner strategies~\citep{wang2024zsc}. GAMMA~\citep{liang2024learning} learns generative models over policy populations, encoding trajectories into latent representations. TALENTS similarly trains generative models but utilizes a sequential architecture to learn long-term strategy representations and performs controlled sampling over these strategy clusters.

\section{Team Adaptation via No-Regret Strategies}

In the following, we introduce TALENTS, a method for training a cooperator agent on a diverse set of strategies and behaviors, enabling it to adapt to unseen teammates.
Given offline trajectories exhibiting various strategies (see Sec.~\ref{sec:vae}), we use a variational autoencoder (VAE)~\citep{kingma2013auto} to learn a latent space of strategy features, then use unsupervised clustering to segment the space into discrete clusters representing unique high-level strategies. Each cluster corresponds to a hidden parameter in a HiP-MDP~\citep{doshi2016hidden} formulation, which we can optimize over by training a best response agent (see Sec.~\ref{sec:agent}).
This is done by generating strategic partners through sampling cluster means and conditioning the cooperator on these means for strategy-dependent responses. 
At test time, we use the fixed share algorithm to select appropriate latent means, enabling adaptation to novel partners (see Sec.~\ref{sec:adaptation}). 

\subsection{Strategy Learning}
\label{sec:vae}
We form a latent strategy space using offline trajectories from agent-agent rollouts (see Sec.~\ref{sec:agentagent}). This dataset, $\mathcal{D}_{traj}={\tau_i}_{i=1}^N$, is composed of observation-action pairs, $\{o^{(i)}_t,a^{(i)}_t\}_{t=0}^{T}, i\in\{1,2\}$, and contains a diverse set of agent strategies and behaviors. We then train a VAE, with encoder $q_{\phi}(z|\tau)$ and sequential decoder network $p_{\theta}(a_{t:t+H}|z, o_t)$, architecture that follows from~\citet{zintgraf2021deep}. As it is intractable to directly maximize the objective of the VAE, we instead optimize the Evidence Lower Bound (ELBO):
\begin{figure}
    \centering
    \includegraphics[width=1\linewidth]{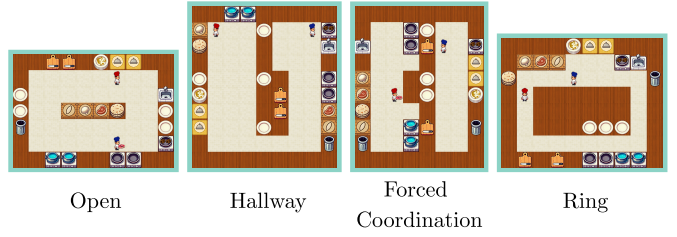}
    \caption{The four Overcooked layouts used in experiments.}
    \label{fig:game_layouts}
\end{figure}

\begin{align*}
\mathcal{L}(\theta,\phi;\tau)
  &= \mathbb{E}_{z\sim q_{\phi}(z\,|\,\tau_{t-h:t})}\bigl[\,
       \log p_{\theta}(a_{t:t+H}\mid z,\,o_t)
     \bigr] \\
  &\quad - \beta\;D_{\mathrm{KL}}\!\bigl(q_{\phi}(z\,|\,\tau_{t-h:t}) \,\|\, p(z)\bigr).
\end{align*}
We then perform K-means clustering with silhouette analysis~\cite{rousseeuw1987silhouettes} to partition the latent space into clusters representing unique agent types for training our cooperator agent.

\begin{algorithm}
\caption{Online Adaptation to Novel Partners with Fixed-Share}
\label{alg:fixedshare}
\begin{algorithmic}[1]
\STATE \textbf{Input:} Latent clusters $\{\mathcal{N}(\mu_1,\sigma_1^2),\ldots, \mathcal{N}(\mu_K,\sigma_K^2)\}$, policy $\pi_{\theta}(\cdot|o, c)$, VAE decoder $p_{\theta}$, switching parameter $\alpha \in (0, 1)$, learning rate $\eta > 0$
\STATE Initialize cluster weight vector $w^1 = (1/K, \ldots, 1/K)$ given $K$ experts
\FOR{$t=1,2,\ldots$}
    \STATE Observe state $o_t$
    \FOR{each cluster $c \in \{1,\ldots, K\}$}
        \STATE Sample latent strategy $z_c \sim \mathcal{N}(\mu_c, \sigma_c^2 I)$
        \STATE Predict partner action $\hat{a}^c_t = \arg\max_a p_{\theta}(a|z_c, o_t)$
    \ENDFOR
  \STATE Compute leading expert $c^* = \arg\max_c w^t_c$, action-biased policy $\pi_{\theta}^{\text{biased}}(\cdot|o_t, c^*)=\pi_\theta(\cdot|o_t)+\mathbf{b}(c^*)$
    \STATE Execute cooperator action $a_t \sim \pi_{\theta}^{\text{biased}}(\cdot|o_t, c^*)$ 
    \STATE Observe partner action $a^p_t$
   \STATE Compute expert losses: $\ell^t_c = -\log p_{\theta}(a^p_t|z_c, o_t)$ 
    \STATE Update pre-sharing weights: $\tilde{w}^{t+1}_c = w^t_c \exp(-\eta \ell^t_c)$
    \STATE Normalize: $\tilde{w}^{t+1} = \tilde{w}^{t+1} / \sum_{c=1}^K \tilde{w}^{t+1}_c$
    \STATE Apply fixed-share update: $w^{t+1}_c = (1-\alpha)\tilde{w}^{t+1}_c + \alpha \sum_{j=1}^K \tilde{w}^{t+1}_j/K$
\ENDFOR
\end{algorithmic}
\end{algorithm}

\subsection{Learning a Strategy-Conditioned Cooperator Agent}
\label{sec:agent}
To train our agent, we leverage the generative ability of the VAE in conjunction with the partitioned strategy clusters.
We first fit Gaussian distributions over each strategy cluster, then retrieve actions for each agent type by decoding samples from each cluster's latent Gaussian, conditioned on the observed environment state. 
The priority-based sampling technique proposed in \citet{zhao2023maximum} is adopted during training to select clusters to sample from, with different clusters representing different types of partners.
We learn a bias vector that corresponds to the cooperator's available actions. This vector modifies the actor network's output logits, explicitly promoting or discouraging specific actions based on the partner type. This approach drives the cooperator to develop distinct conventions tailored to each partner's behavior. We train the cooperator agent using independent PPO~\citep{de2020independent,schulman2017proximal}.

\subsection{Online Adaptation to Novel Partners}
\label{sec:adaptation}
The trained cooperator can best respond to any partner strategy in the VAE distribution given knowledge of the partner's latent strategy, thus requiring online partner type inference.
We employ regret minimization with a fixed-share~\cite{herbster1998tracking} algorithm variant, treating each strategy cluster as an ``expert'' (Algorithm~\ref{alg:fixedshare}).
Fixed-share differs from static no-regret methods (Hedge~\citep{freund1997decision}, FTRL~\citep{mcmahan2011follow}) by minimizing tracking regret, accounting for non-stationary test-time policies that change strategy due to partner influence or continuous adaptation.
This provides regret bounds ensuring that encountering truly novel strategies only degrades performance to the best-response of the closest observed training strategy.

\section{Experiments}

We evaluate our approach in a modified version of the Overcooked-AI environment across agent-agent and human-agent gameplay. 
In this domain, we evaluate across four layouts (see Fig.~\ref{fig:game_layouts}).
We utilize four environments for agent-agent teams, and three for human-agent teams. 
At their core, the selected layouts are conceptually consistent with those proposed in~\citep{carroll2019utility}, but demand more effective team coordination through additional task complexity (order list, more dish types) and timing constraints (expiring orders, burnt dish risk).

\subsection{Agent-Agent Zero-Shot Coordination}
\label{sec:agentagent}

We begin by evaluating our agent in an agent-agent setting across three diverse partner populations: Fictitious Co-Play (FCP)~\citep{strouse2021collaborating}, Maximum Entropy Population (MEP)~\citep{zhao2023maximum}, and Behavior Preference (BP) agents~\citep{wang2024zsc}. 
We use BP as a baseline following the finding from~\citet{wang2024zsc} that BP populations exhibit diversity that more closely resembles that of human teams.
Each BP policy is trained using a linear combination of reward shaping terms to encourage the emergence of distinct strategies, for example, a term that explicitly rewards rice boiling or one that rewards dishwashing.
We train 12 policies for each agent population, keeping three training checkpoints from each to form a 36-policy population.
To train a TALENTS agent, we first generate trajectory data by selecting agents from all populations and playing games between all pairs of agents.
Using the resulting rollouts, we train the VAE to predict the next action of each agent. 
We then cluster the learned latent space as described in Section~\ref{sec:vae}. 
Subsequently, we compare TALENTS against two baselines: a population-trained best-response cooperator and a GAMMA~\citep{liang2024learning} cooperator, which, like our method, incorporates a generative model during training. 
To compare the adaptability of the agents when paired with novel partners, we utilize a held-out set of Behavior Preference agents (see Table~\ref{tab:main_results}).
With this experiment, our objective was to assess whether the strategy-specific partner modeling and cooperator conditioning of TALENTS leads to greater robustness and coordination performance across diverse agent populations.

Across most layouts, TALENTS achieves the highest reward among all evaluated methods, regardless of the training population. 
However, in the \textit{Forced Coordination} layout, the population best-response agent outperforms both TALENTS and GAMMA.
We hypothesize that this is due to the rigid division of tasks required to succeed in that layout.
TALENTS and GAMMA are specifically designed to model and adapt to a broad distribution of partner behaviors, including suboptimal or uncooperative teammates, which may reduce their effectiveness in such structured scenarios where precise role adherence is critical.

\begin{figure}[t]
    \centering
    \includegraphics[width=0.44\textwidth]{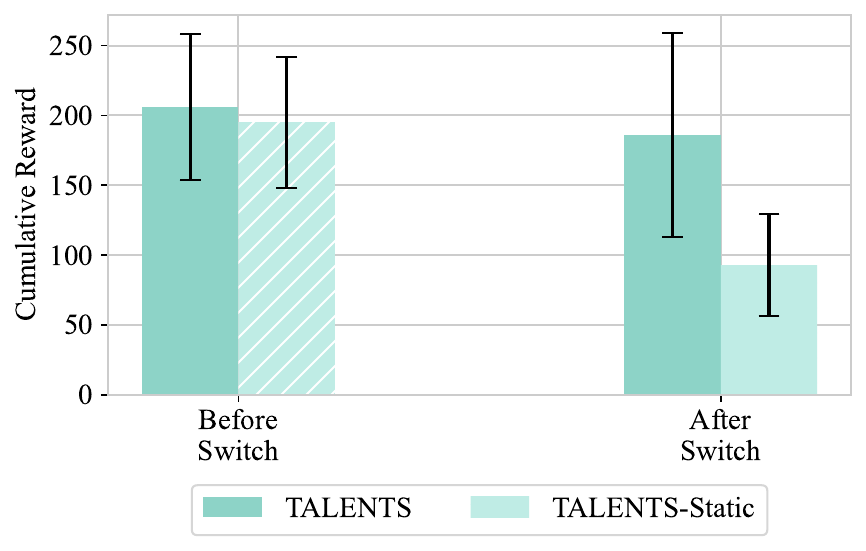}
    \caption{Accumulated reward in the first and second halves of the episode when the partner policy is switched midway through the episode. Results are averaged across evaluations performed in the 4 layouts depicted in Fig~\ref{fig:game_layouts}. }
    \label{fig:ablateplot}
\end{figure}

Furthermore we perform a set of ablations to assess the importance of the tracking-regret minimization component in which the fixed-share algorithm is replaced with a static-regret minimizer.
To this end, we retain the exponential weight update rule but remove the weight-sharing mechanism that allows adaptation over time, yielding a baseline equivalent to a standard Hedge algorithm~\citep{freund1997decision}. 
Subsequently, we evaluate our cooperator in a setting where the partner policy is switched midway through the episode to a different held-out Behavior Preference policy that behaves according to a different strategy.
As shown in Fig.~\ref{fig:ablateplot}, TALENTS-Static fails to update its belief in response to the new partner, due to it having an irrecoverably strong bias toward the original partner strategy. This results in a notable drop in reward in the second half of the episode, highlighting the importance of dynamic partner modeling for maintaining robust coordination.

\begin{table}[]
  \caption{Resulting mean scores of games played with a held-out set of 12 diverse behavior-preferenced SP agents. Scores are averaged over 3 sets of policies, trained with FCP, MEP, and BP populations.}
  \centering
  \footnotesize
  \begin{tabular}{lcccc|c}
    \toprule
    \textbf{Agent} & \textbf{Open} & \textbf{Hallway} & \textbf{\makecell{Forced\\Coord}} & \textbf{Ring} & \textbf{Overall} \\
    \midrule
      TALENTS & $\mathbf{461.02}^{*}$ & $\mathbf{688.48}^{*}$ & $28.26$             & $\mathbf{547.22}^{*}$ & $\mathbf{431.25}^*$ \\
      GAMMA   & $418.41$          & $376.82$          & $41.10$             & $454.48$           & $322.70$ \\
      BR      & $386.96$          & $231.72$          & $\mathbf{69.70}^{*}$     & $90.34$            & $194.68$ \\
    \bottomrule
  \end{tabular}
 
  \vspace{4pt}
  \footnotesize
  \textsuperscript{\textbf{*}} $p<0.05$, paired t-test, Bonferroni-corrected (significantly better than second-best agent) 
  \label{tab:main_results}
\end{table}

\begin{figure*}[t] 
\centering 
\includegraphics[width=1\textwidth]{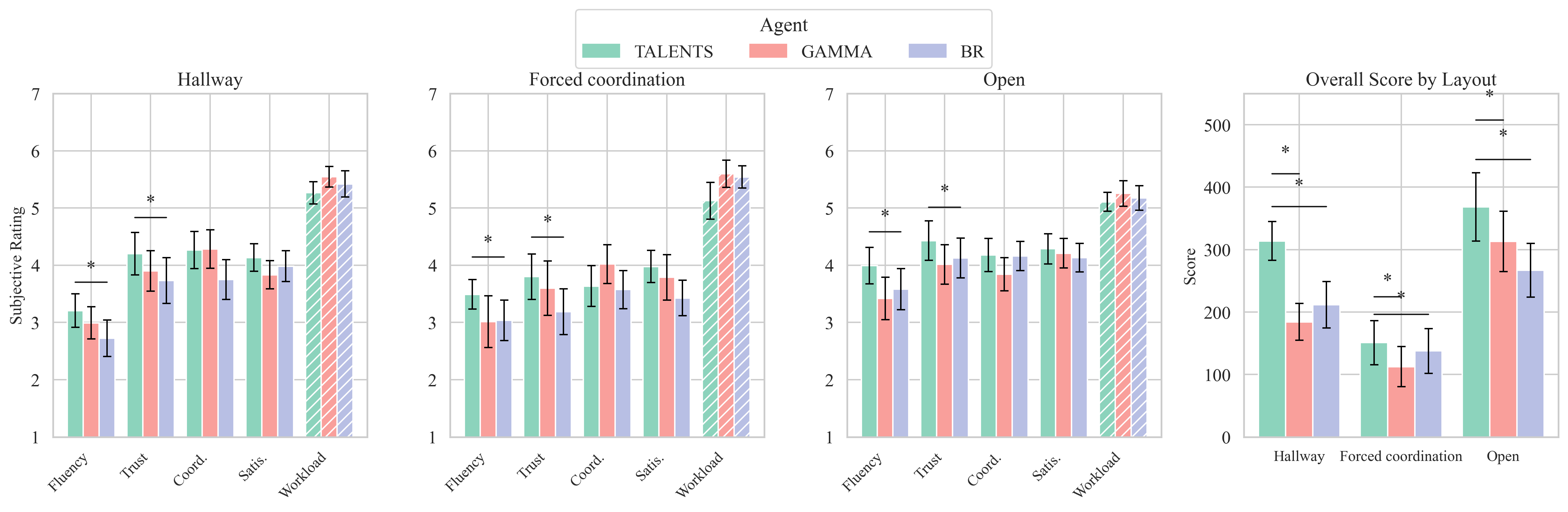} \caption{Human-agent teamwork evaluation results are measured using team score and participants' subjective ratings of their agent teammates. Perceived workload (shaded) is better if lower. Asterisks represent statistically significant differences across agents. Error bars are one standard error from the mean.} \label{fig:hat} \end{figure*}

\subsection{Human-Agent Zero-Shot Coordination} \label{sec:IRBDetails} We evaluate the TALENTS agent against state-of-the-art baselines with human participants recruited from Prolific and Cloud Research. Each participant completed three four-minute rounds with different agent partners and received \$7 base payment plus up to \$3 performance bonus. 
During the study, the agents' action frequency was limited to ensure their input speed matched that of average human players.

\subsubsection{Experiment Design} \label{sec:IRBSurveys} We employed a mixed design with agent type as the within-subject variable (3 levels: ours, GAMMA, Population Best Response) and map layout as the between-subject variable (3 levels: hallway, open, forced coordination). After each round, participants completed questionnaires measuring workload (NASA-TLX~\citep{Hart1988-mp}), team fluency, trust, coordination, and satisfaction on 7-point Likert scales.

\subsubsection{Results} We collected data from 119 participants, filtering out inactive participants and incorrect trap question responses. As shown in Fig~\ref{fig:hat}, our agent significantly outperforms both baselines. Mixed ANOVA tests show significant main effects for team score ($F(2,166)=5.76, p=.003$), with our agent achieving significantly higher scores than both baselines in post-hoc comparisons ($p < .05$). Our agent also receives higher subjective ratings in team fluency ($F(2,122)=4.31, p=.02$) and perceived trust ($F(2,122)=3.23, p=.04$) than the BR baseline ($ps < .05$).

\section{Conclusion}

In this work, we propose TALENTS, a novel ZSC agent that leverages knowledge of agent types to effectively collaborate with and adapt to teammates. Our method employs a variational autoencoder to learn a latent strategy space, which is then partitioned using unsupervised clustering to identify distinct partner types. The VAE's generative ability then enables teammate generation and conditioning during cooperator training. 
During testing, we use fixed-share for partner type inference and adaptation in online settings, addressing a key limitation of existing methods by enabling continuous strategy adjustment based on observed teammate behavior.
Through agent-agent and human-agent experiments in the Overcooked testbed, we demonstrate that TALENTS adapts to partner strategies and achieves higher performance than state-of-the-art population-based methods (FCP, MEP, BP) and a generative agent method (GAMMA). In human-agent experiments, participants prefer TALENTS over baseline methods in fluency and trust metrics.
In the future, TALENTS can be extended to teams with more than two players or explore applying the framework to train agents that not only adapt to their teammates but also influence them toward more optimal strategies.

\section{Acknowledgments}
This research is supported by the Defense Advanced Research Projects Agency EMHAT program, agreement No. HR00112490409.

\bibliographystyle{plainnat}
\bibliography{references}

\end{document}